\documentclass[letterpapper, 10pt, conference]{ieeeconf}
\IEEEoverridecommandlockouts    
\usepackage[T1]{fontenc}        
\usepackage[utf8]{inputenc}     
\usepackage[english]{babel}     
\usepackage[plain]{fancyref}    
\usepackage{graphicx}
\usepackage{textcomp}
\usepackage{float}              
\usepackage{svg}
\usepackage{booktabs}
\usepackage{hyperref}           
\hypersetup{colorlinks=true, linkcolor=black, citecolor=black}    
\usepackage{csquotes} 
\usepackage[super]{nth}
\usepackage{hypcap} 
\usepackage{amsmath,amsfonts,amssymb}
\usepackage{siunitx} 
\usepackage[
backend=biber,
style=numeric,
sorting=none 
]{biblatex}
\usepackage{tikz}			
\usetikzlibrary{
	calc,		
	decorations.markings,
	arrows,
	arrows.meta,
	optics,
	external,
	shapes.geometric,
	patterns,
}
\usepackage{pgfplots}
\usepackage{xcolor}			
\newcommand{\comment}[1]{}

\usepackage{dblfloatfix}

\title{\LARGE \bf
The Software Stack That Won the \\Formula Student Driverless Competition
}

\author{Andres Alvarez$^{1}$, Nico Denner$^{2}$, Zhe Feng$^{2}$, David Fischer$^{1}$, Yang Gao$^{1}$,\\ Lukas Harsch$^{2}$, Sebastian Herz$^{1}$,
Nick Le Large$^{1}$, Bach Nguyen$^{1}$, Carlos Rosero$^{1}$,\\Simon Schaefer$^{1}{^{\dagger}}$, Alexander Terletskiy$^{1}$, Luca Wahl$^{1}$, Shaoxiang Wang$^{1}$, Jonona Yakupova$^{1}$, Haocen Yu$^{2}$
\thanks{$^{1}$ Author and Researcher}%
\thanks{$^{2}$ Researcher}%
\thanks{$^{1}$ $^{2}$ Karlsruhe Institute of Technology and KA-RaceIng e.V., \textit{firstname.lastname@ka-raceing.de}}%
\thanks{$^{\dagger}$ Corresponding author, \textit{simon.schaefer@ka-raceing.de}}%
}

\begin{document}

\hyphenation{LiDAR}
\hyphenation{image}

\hyphenpenalty=300

\maketitle
\pagestyle{plain}


\begin{abstract}
This report describes our approach to design and evaluate a software stack for a race car capable of achieving competitive driving performance in the different disciplines of the Formula Student Driverless.
By using a 360° LiDAR and optionally three cameras, we reliably recognize the plastic cones that mark the track boundaries at distances of around \SI{35}{\meter}, enabling us to drive at the physical limits of the car. Using a GraphSLAM algorithm, we are able to map these cones with a root-mean-square error of less than \SI{15}{\cm} while driving at speeds of over \SI{70}{\km \per \hour} on a narrow track. The high-precision map is used in the trajectory planning to detect the lane boundaries using Delaunay triangulation and a parametric cubic spline. We calculate an optimized trajectory using a minimum curvature approach together with a GGS-diagram that takes the aerodynamics at different velocities into account. To track the target path with accelerations of up to \SI{1.6}{\g}, the control system is split into a PI controller for longitudinal control and model predictive controller for lateral control. Additionally, a low-level optimal control allocation is used.
The software is realized in ROS C++ and tested in a custom simulation, as well as on the actual race track.
\end{abstract}
 
\section{INTRODUCTION}

In the Formula Student competitions, based on extensive rules and guidelines similar to Formula SAE, student teams throughout the world design and manufacture an open-wheel, single-seater race car. Originally consisting of only combustion vehicles, the competition was since extended with an electric category, and starting in 2017, with an autonomous category (\textit{Formula Student Driverless}) as well. Points are awarded for various aspects, the most substantial of which are the quality of the engineering design as well as the on-track performance. One of the most technically challenging disciplines, Autocross, consists of an unknown, closed-loop and narrow track of around \SIrange{200}{300}{m} length 
outlined by yellow and blue plastic cones, which must be completed as quickly as possible without hitting any of the cones. While on track, any interaction with, or remote control of the vehicle, is forbidden. 

\begin{figure}
\begin{center}
\includegraphics[width=0.5\textwidth]{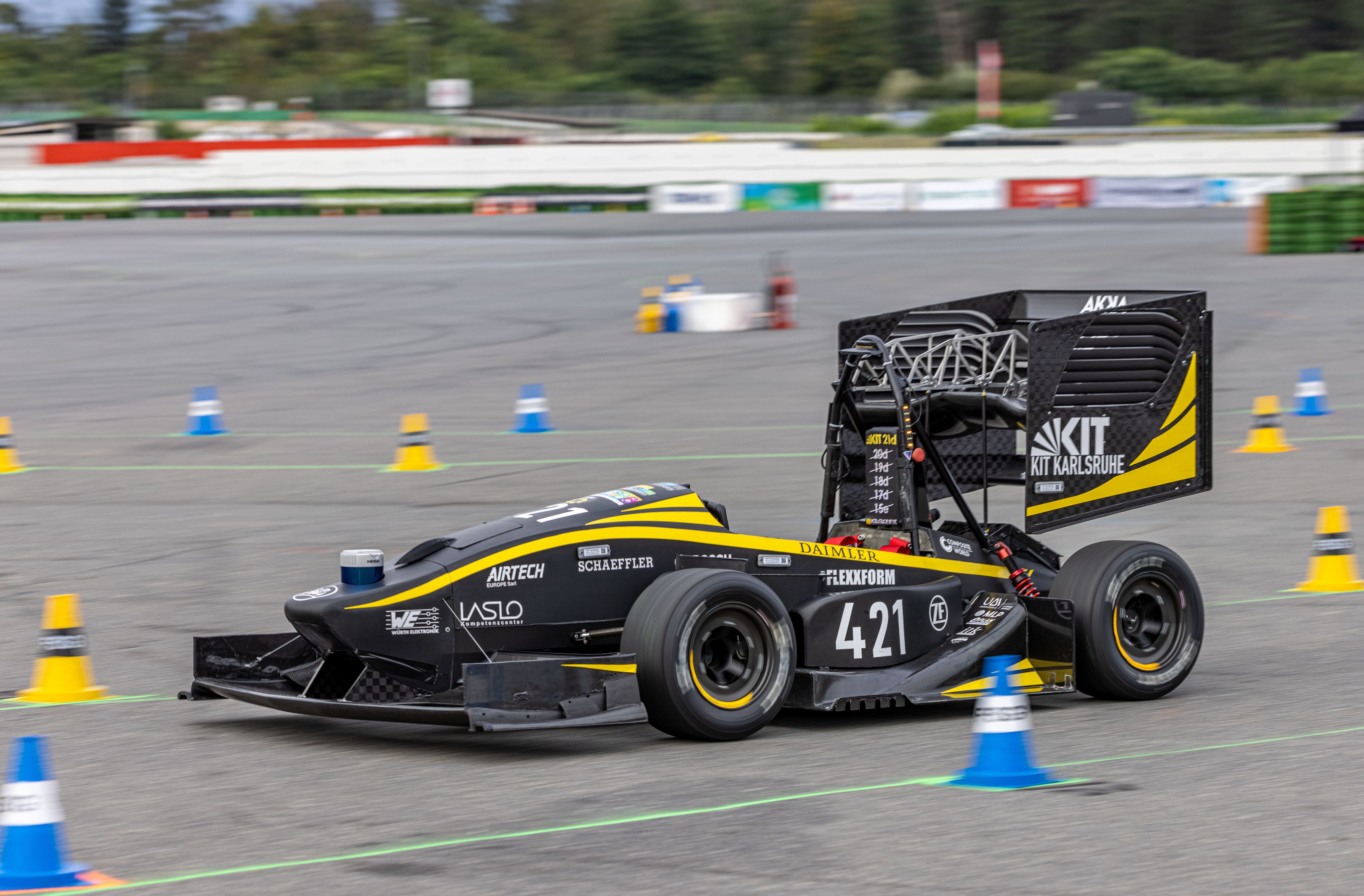}
\caption{The KIT21d driving at Formula Student Germany 2021. Photo credit: FSG Partenfelder.}\label{fig:kit21d}
\end{center}
\end{figure}

Founded in 2006 by students of the Karlsruhe Institute of Technology, the team KA-RaceIng 
developed their \nth{5} autonomous car for the 2021 competition. The \textit{KIT21d} is shown in Figure \ref{fig:kit21d}. It features a carbon fiber-reinforced polymer (CFRP) chassis that is equipped with four electric motors with a maximum power of \SI{80}{kW} in total, a \SI{470}{V} battery with a capacity of \SI{5.2}{kWh}, and weighs \SI{214}{kg}. 

\begin{figure*}
\begin{center}
\includegraphics[width=0.99\textwidth]{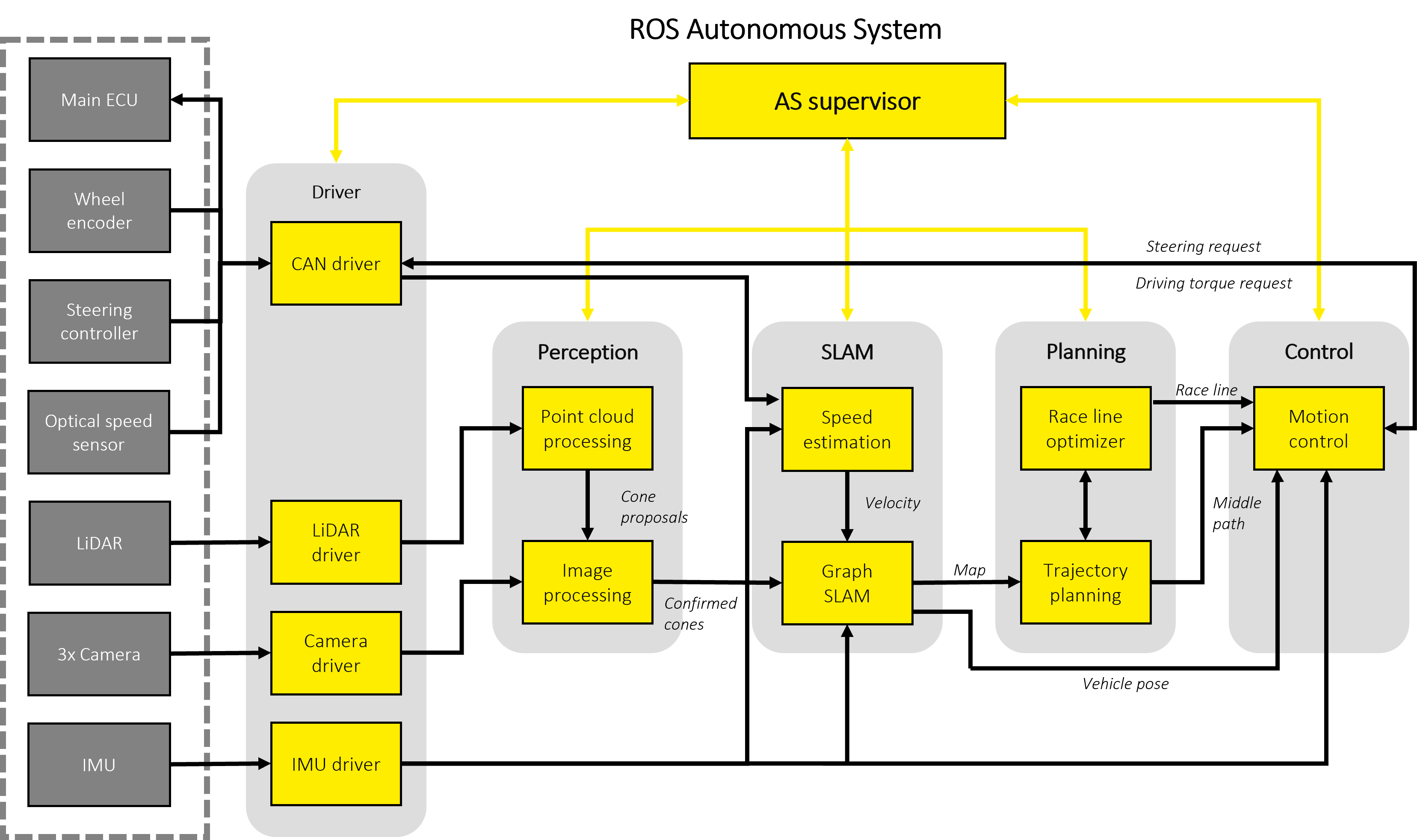}
\caption{System Overview}\label{architecture_diagram}
\end{center}
\end{figure*}
 
\section{Design Goals}
After finishing \nth{2} overall three years in a row at Formula Student Germany between 2017 and 2019, our main goal for 2020/2021 was a \nth{1} place overall at all events. In the Autonomous System, we focused on two points to achieve this goal.

\textbf{Increased robustness in localization and path-planning}
The analysis of data collected during the test days and events showed that our car was regularly on the verge of taking a wrong turn. The planned trajectory was corrected only in the last second, meaning we drove at the absolute limit. To drive any faster without making trade-offs in safety, we needed a correct trajectory much further ahead. To achieve this, improvements were needed in the first three modules of the autonomous pipeline:
\begin{enumerate}
\item Perception: In 2019, cones were first detected at a distance of approximately \SI{30}{\meter}, with the median lying at around \SI{20}{\meter}. We set the goal to increase both figures by at least \SI{10}{\meter}, while maintaining a false-positive rate near zero.
\item SLAM: To complete the \SI{40}{\meter} perception range goal, SLAM needed to be able to handle the increased number of landmarks by utilizing a parallelized architecture.
\item Planning: The generation of a correct path depends on interpreting the mapped landmarks correctly. Our goal this year was to evaluate new algorithms and compare them to last year's method in terms of accuracy in difficult situations and computation time.
\end{enumerate}
\textbf{Increasing average speed}
\begin{enumerate}
\item On straights: To increase acceleration, we set the goal of implementing a traction control system.

\item In corners: To use as much of the track width as possible, the precision of the pose estimation and path tracking had to be increased. Additionally, torque vectoring and active yaw rate control were required to ensure stability in highly dynamic situations.
\end{enumerate}

\section{System Overview}
The autonomous system software runs centrally on a multi-core x86
processing unit (Autonomous Computing Unit, ACU), which provides the necessary computational power to run our autonomous system in real time. If cameras are used, this x86 CPU is complemented by the Coral Edge TPU machine learning co-processor used for running an image classification neural network. The sensors shown in figure \ref{architecture_diagram} are connected directly to the ACU via USB3, Ethernet or CAN. All actuation values are sent via CAN directly to the Electronics Control Unit (ECU), which manages the electrical system of the car and continuously performs safety checks on the complete system. 
Figure \ref{architecture_diagram} provides a high-level overview of the communication in our autonomous system. The system is implemented using the Robot Operating System (ROS) framework in the Melodic Morenia release. Most components are implemented in C++, except for some smaller modules realized in Python. The central processing pipeline starts with perception. The cones detected by the perception system are processed by the SLAM algorithm, which localizes the vehicle and builds a map of its surroundings. On this map, the target trajectory is planned and then realized by the motion control system. This whole process is constantly monitored by the supervisor node that performs health and sanity checks of the other nodes to ensure a safe drive. Additionally, our \textit{Simulation} is capable of testing all the path planning and control parts of the pipeline outside of the car, aiding us in fine-tuning the system and reducing necessary test time.
 
\section{Perception}\label{perception}

The perception system is responsible for recognizing the position and color of the cones that define the race track. The pipeline takes advantage of the precise location information provided by the LiDAR. Additionally, rich semantic information provided by the cameras can be included if necessary.

\comment{
\begin{figure*}[b]
\begin{center}
\def\svgscale{0.65}
\includesvg[inkscapelatex=false]{images/perception_overview.svg}
\caption{The Perception Pipeline}\ \label{perception_archictecture}
\end{center}
\end{figure*}
}

\subsection{LiDAR System}
The KIT21d uses one Hesai Pandar40P, a mechanically rotating LiDAR operating at \SI{10}{\hertz}, to acquire precise position estimates of the cones. Because of its placement on top of the vehicle's nose, the field of view is limited to approximately 240°. 

During the pre-processing, the raw point cloud is reduced by discarding irrelevant points that are too far away to be on the race track or originate from the monocoque, resulting in a 60\% reduction. Then, using a ground plane estimation algorithm described in \cite{Himmelsbach2010FastSO} that takes into account the slope between successive points, all data points assumed to be on the ground plane are removed from the cloud, reducing the point cloud by a further 30\%.
Since we use the LiDAR at a frequency of \SI{10}{\hertz}, the distortion caused by the movement while scanning can not be neglected and is corrected using odometry and acceleration measurements.
To extract cone positions from the remaining points, a 2D euclidean clustering and a neighbourhood filtering are performed. Clusters that do not fulfil the size or layout requirements of Formula Student race track cones are filtered out. These remaining cluster centroids are assumed to be cones. This whole process takes approximately \SI{32}{\milli \second}.

\subsection{Optional Sensor Fusion and Classification System}\label{section:Fusion}
The LiDAR-only pipeline achieves a visual range of at least \SI{35}{\meter} and a false positive rate of less than 1\% for data recorded on our test track at the campus. A further improvement in robustness and reliability is provided by the optional computer vision module. When it is enabled, the centroids provided by the point cloud processing are considered to be cone proposals. Using corresponding feature points, these 3D points in the LiDAR’s local coordinate system are projected into the 2D camera image space with a pinhole camera model, whereby a bounding box is created in the image space. The extrinsic parameters (rotation and translation) are obtained via a non-linear fit while the intrinsic and distortion parameters are obtained via the Autoware-AI checkerboard-based calibration toolkit \cite{kato2018autoware}. The created bounding boxes are then passed on to the image classification system.

This system is composed of three forward-facing RGB mono cameras with lenses with different focal lengths and placed at different angles, providing a combined field of view of approximately 180°.
The cropped image segments from the camera image bounding boxes are classified by an efficient convolutional neural network, leading to the recognition of the type of object contained in them: \textit{blue cone}, \textit{yellow cone}, \textit{orange cone}, and \textit{not a cone}. 
We also achieve very high energy efficiency by not using a GPU to run our neural network. Instead, we utilize the Coral Edge TPU, meaning we use \SI{0.5}{Watt} for each Tera-operation per second (TOPS) and \SI{4}{Watt} in peak, while experiencing no significant disadvantage regarding the network's inference speed.
The network is trained on self-accumulated data from test runs (in different lighting and weather conditions, with data augmentation), carried out in the lead up to the formula student competitions. 
When the perception pipeline is also utilizing the vision module, it is comparable to a Faster R-CNN \cite{DBLP:journals/corr/RenHG015}, but instead of using a Region Proposal Network, the LiDAR clusters serve as the region proposals. 

Even though this module enables a median detection range of \SI{40}{\meter}, it is only a \SI{5}{\meter} improvement over the LiDAR-only pipeline, which already had met the design goal. Therefore, this module was not used in the 2021 competitions, as the increasing latency, complexity and computing cost necessary for it outweighed its benefit. As Formula Student is also an engineering design competition and not just a race, showing this modular pipeline and evaluating the approaches scored us valuable extra points. 

\section{SLAM}

Simultaneous Localization and Mapping (SLAM) is the problem in which a vehicle tries to build a map of landmarks and to locate itself in this map at the same time. A classical SLAM approach often has two parts: front-end and back-end. In the front-end, necessary data like odometry and visual observations are acquired and fed into a mathematical model. In the back-end, this model is optimized to produce the most precise map and vehicle pose possible.

\subsection{Velocity Estimation}
In the simplest form, the current state of the vehicle (velocity and orientation) is determined using an IMU and wheel speeds. These measurements, called odometry data, are complemented by an optical ground speed sensor. Additionally, we have designed and trained a \textit{Recurrent Kalman Network} \cite{rkn2019}, which is basically a Kalman filter where the measurement model and dynamic model are obtained using deep learning, with the optical ground speed sensor used as a ground truth. This network has a mean error of about \SI{0.9}{\km\per\hour}, which is about 70\% of the error obtained by direct velocity calculations from wheel speeds. 

\subsection{Data Association}
In the front-end, the Data Association (DA) has the task to match each new observation to one of the previously mapped landmarks. If no association is possible, the landmark is assumed to be new. A well-working DA is the basis of a good SLAM, as false positive and false negative associations quickly lead to a reduction in accuracy of the estimation.

A straightforward DA algorithm called k-nearest neighbors (kNN) is employed. In kNN, for each observation, \textit{k}\footnote{Setting \textit{k} to 10 worked well for us.} landmarks nearest to this observation are queried from the map. The landmark map is built as a kd-tree.
Since the measurement of observations and the estimated cone position carries noise, instead of the usual Euclidian distance, the Mahalanobis distance (that takes the uncertainty of the estimated position into account) is used to find the nearest landmarks. From the \textit{k} found landmarks for an observation, the closest one that is not yet associated with another observation is matched. In comparison to a more complex algorithm with exponential runtime complexity, JCBB \cite{jcbb}, kNN has similar performance in most cases while being much faster at linear-logarithmic runtime complexity.

\subsection{GraphSLAM}
\begin{figure}[H]
\begin{center}
    \includegraphics[width=0.45\textwidth]{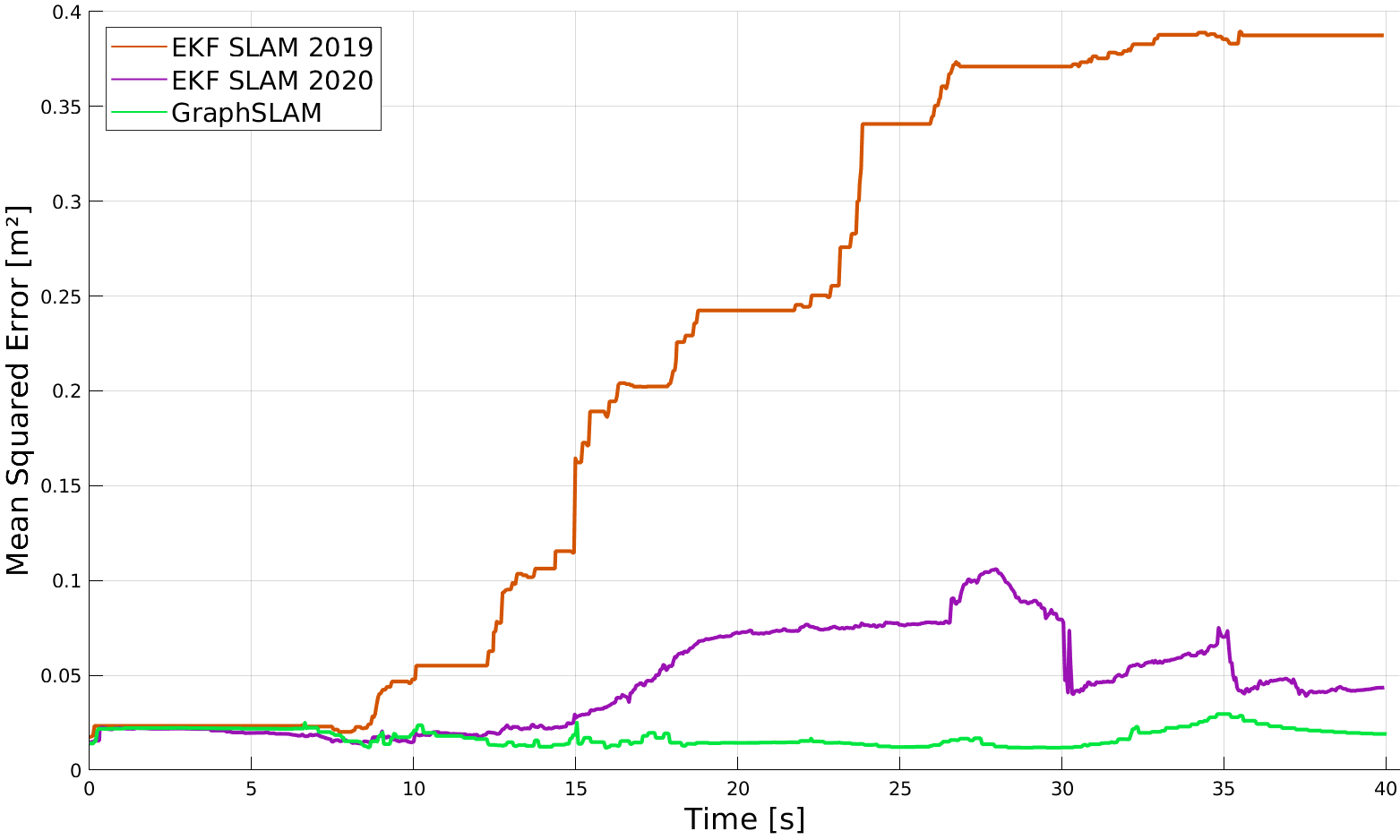}
    \caption{Autocross mean squared error of all cone positions for the GraphSLAM, the newly developed EKF SLAM, and the EKF SLAM from season 2019}
    \label{fig:SLAM_Error}
\end{center}
\end{figure}

A graph-based approach is implemented using the g2o library \cite{kuemmerle11g2o}. In GraphSLAM, the SLAM problem is represented in an overdetermined linear system of equations using a graph structure. Landmarks and vehicle poses in the graph are variables. State measurements are seen as constraints connecting two consecutive vehicle poses, while landmarks measurements are used to connect landmark measurements to a vehicle pose. Solving the resulting system of equations means finding the configuration of vehicle poses and landmarks that best fits all available constraints. Being built on a parallelized architecture, the algorithm is able to perform in real time. The graph is built continuously, while being optimized in the background as often as possible. Odometry data is used to get a pose estimate between two optimizations. Compared to our self-developed Extended Kalman Filter (EKF) SLAM based on \cite{thrun2005probabilistic}, GraphSLAM produces a more accurate map and vehicle poses while still running fast enough within the autonomous system. For a reference track the EKF SLAM has a maximum CPU usage of $26.14\%$ and the GraphSLAM  reaches $43.93\%$. However, the final mean squared error of all cone positions of the GraphSLAM is \SI{0.0189}{\meter \squared} and \SI{0.0436}{\meter \squared} for the EKF SLAM. The ground truth map for the calculation of the error was acquired with high-precision DGPS measurements. Figure \ref{fig:SLAM_Error} depicts the results of the error comparison. This comparison of the two SLAM algorithms in the context of Formula Student has been published in \cite{lelarge}.

\section{Trajectory Planning}

The goal of this year's trajectory planning was to refine the algorithm based on the Delaunay Triangulation developed last year and to introduce an optimization module capable of producing optimal trajectories off- and online.

Our current pipeline consists of the following steps: the landmarks received from SLAM are used as vertices to compute the Delaunay triangles. The triangulation is carried out with the divide and conquer algorithm which is proven to be the fastest Delaunay Triangulation generation technique \cite{delaunay}. 

Next, using the properties of the vertices and edges of the triangles as well as their relative position (distance and angle), triangles can be filtered and sorted. At the same time, the boundaries of the track and the edges connecting them are saved. This allows us to treat the middle points of these connecting edges as the control points for the interpolation. 

The obtained control points are smoothed and then connected by interpolation with a parametric cubic spline. Parameters of the splines are calculated using the self-written Tridiagonal-Matrix-Algorithm (TDMA), which is faster than a sparse-matrix solver library. The TDMA is a simplified form of Gaussian elimination, which solves tridiagonal systems of equations \cite{tdma}. One of the significant advantages of using the parametric cubic spline is that at each point the curvature, heading and arc length can be easily calculated and used later in velocity planning and in control.

The velocity is planned based on the GGS-diagram that incorporates our vehicle model parameters, like tire friction, drag and lift coefficients, power and acceleration limits. 

\begin{figure}[H]
\begin{center}
    \includegraphics[width=0.4\textwidth, angle=90]{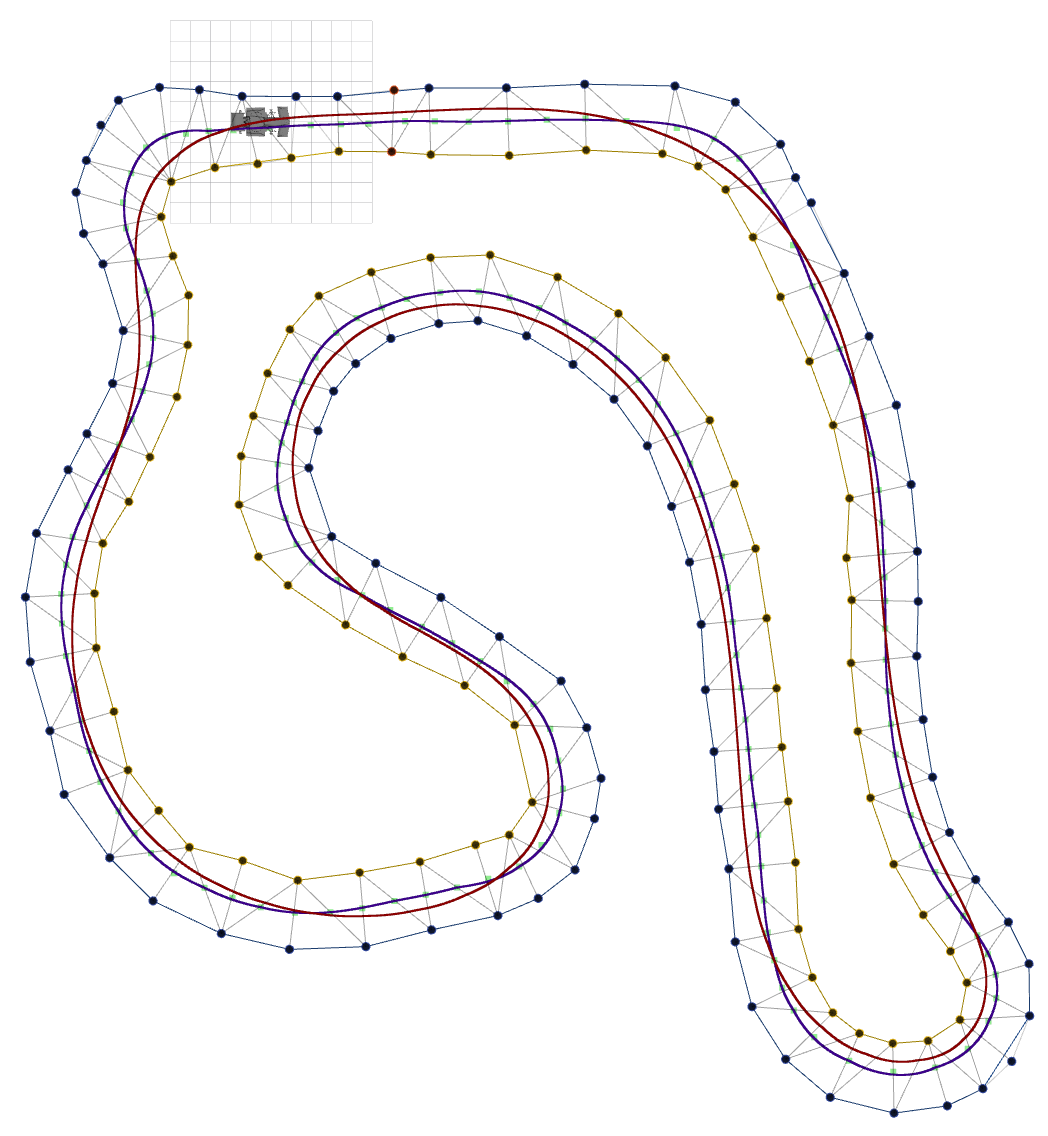}
    \caption{Visualization of trajectory planning on the Formula Student Spain track. Blue line: middle path, red line: optimized race line.}
    \label{fig:mincurv}
\end{center}
\end{figure}

The newest addition to our pipeline this year is the minimum curvature trajectory optimization \cite{mincurv} as shown in figure \ref{fig:mincurv}, which runs in parallel to the middle line generation. This algorithm minimizes the total curvature along the track and produces smoother trajectories that provide more stability for the controller and faster lap times. We were able to adjust the method to operate online on unclosed tracks. Even on narrow race tracks, a reduction in lap time of more than $10 \% $  can be achieved.

\section{Motion Control} \label{sec: Control}
To fulfill the goal of trajectory tracking, we chose a two-layer motion control pipeline based on a high- and a low-level controller. The high-level controller uses a decoupled control structure based on a model predictive controller (MPC) for lateral vehicle control and a 2 DoF PI controller for velocity tracking, as described in \cite{nekkah2020autonomous}. The low-level controller consists of an active yaw rate control using optimal control allocation (OCA) methods to drive the car at its handling limits.
\subsection{Lateral MPC}
Using a dynamic bicycle model excluding the longitudinal dynamics and linearized around straight-line driving, we get the continuous-time linear model, 
\begin{equation}\label{linear_model}
\dot{\boldsymbol{x}}=\mathbf{A} \boldsymbol{x}+\mathbf{b} u,
\end{equation}

presented in \cite{nekkah2020autonomous}. The states of the system, $\boldsymbol{x}=\left[y, v_{y}, \psi, \dot{\psi}\right]^\intercal$, model the yaw and lateral dynamics of the car with the steering angle as scalar input $u=\delta$.

We formulate a linear time-variant MPC (LTV-MPC) as a linear-quadratic optimization problem in terms of the control input rate $\Delta u$ from which $u$ is obtained by accumulation and using the discretized version of (\ref{linear_model}): 

\begin{subequations}
	\label{eq:LTV_MPC}
	\begin{alignat}{2}
	& \underset{\substack{\Delta u_{1:N}, \\ \boldsymbol{x}_{1:N+1}}}{\text{min}} \
	& & \sum^{N}_{k=1} \left(\Vert \boldsymbol{x}_k-\boldsymbol{x}^{ref}_k \Vert^2_\mathbf{Q} \nonumber + R \Delta u_k^2 \right)   \\
	\label{eq:LTV_MPCa}
	&
	& & \qquad \qquad \qquad + \Vert \boldsymbol{x}_{N+1}-\boldsymbol{x}^{ref}_{N+1} \Vert^2_\mathbf{P} \\
	\label{eq:LTV_MPCb}
	& \text{s.t.}
	& & \boldsymbol{x}_{k+1} = \mathbf{A} \boldsymbol{x}_k + \mathbf{b} u_k \quad k=1,..,N\\
	\label{eq:LTV_MPCc}
	&
	& & u_{k+1} = u_k + \Delta u_k \quad k=1,..,N\\
	\label{eq:LTV_MPCd}
	&
	& & \boldsymbol{x}_1 = \hat{\boldsymbol{x}}\\
	\label{eq:LTV_MPCe} 
	&
	& & u_1 = \hat{u}\\
	\label{eq:LTV_MPCf}
	&
	& & \mathbf{D} \boldsymbol{x}_k + \mathbf{e} u_k + \mathbf{f} \leq \mathbf{0} \quad k=1,..,N+1\\
	\label{eq:LTV_MPCg}
	&
	& & \underline{\boldsymbol{x}} \leq \boldsymbol{x}_k \leq \overline{\boldsymbol{x}} \quad k=1,..,N+1\\
	\label{eq:LTV_MPCh}
	&
	& & \underline{u} \leq u_k \leq \overline{u} \quad k=1,..,N+1 \\
	\label{eq:LTV_MPCi}
	&
	& & \underline{\Delta u} \leq \Delta u_k \leq \overline{\Delta u} \quad k=1,..,N.
	\end{alignat}
\end{subequations}

By largely following the methods presented in  \cite{maciejowski2002predictive}, we can take advantage of the linear system dynamics to express the sequence of system states over the receding horizon merely in terms of the sequence of control input rates. Thus, the state variables are being eliminated from the optimization problem and the system dynamics is satisfied in each step of the optimization horizon. The resulting problem has the form of a constrained Quadratic Program (QP),

\begin{subequations}
	\label{eq:denseQP}
	\begin{alignat}{2}
	\label{eq:denseQPa}
	& \underset{\boldsymbol{\Delta u}_{1:N}}{\text{min}} \quad
	& &  \frac{1}{2} \boldsymbol{\Delta u}^{\intercal}_{1:N} \mathbf{H} \boldsymbol{\Delta u}_{1:N} + \mathbf{g}^{\intercal} \boldsymbol{\Delta u}_{1:N}\\
	\label{eq:denseQPb}
	& \text{s.t.} \quad
	& & \underline{\mathbf{d}} \leq \mathbf{D}\boldsymbol{\Delta u}_{1:N} \leq \overline{\mathbf{d}} \\
	\label{eq:denseQPc}
	& 
	& & \underline{\Delta u} \leq \Delta u_k \leq \overline{\Delta u},
	\end{alignat}
\end{subequations}

where all decision variables have been gathered in a single vector $\Delta \boldsymbol{u}_{1: N}=\left[\Delta u_{1}, \Delta u_{2}, \ldots, \Delta u_{N}\right]^\intercal$. Experience from previous Formula Student competitions suggest that the maximum steering angle is hardly reached when operating the vehicle within its physical limits on representative Formula Student tracks. Furthermore, our minimum-curvature path optimization already considers the maximum geometric curvature that can be achieved by our vehicle. Thus, the constraints of problem (\ref{eq:denseQP}) can be dropped, leading to an unconstrained QP. The solution can be obtained by solving the following system of linear equations by means of the Cholenksy decomposition algorithm of the Eigen C++ library \cite{eigenweb}: 

\begin{equation}\label{qp_solution}
\mathbf{H} \boldsymbol{\Delta} \boldsymbol{u}_{1: N}=-\mathbf{g}.
\end{equation}

This basic MPC controller can be enhanced by means of a steer delay compensation feature that considers the steering actuator dynamics. The latter can be modeled as a second order lag (PT2) element and represented in the state space with the state vector $\boldsymbol{\xi} = \left[\tilde{\delta}, \dot{\tilde{\delta}} \right]^\intercal$ by

\begin{subequations}\label{pt2_state_space_system_1}
    \begin{alignat}{2}
    &\dot{\boldsymbol{\xi}}=\mathbf{A}_{\xi}(T,D)\: \boldsymbol{\xi} + \mathbf{b}_{\xi}(T,D)\:\delta \\
    &\tilde{\delta}=\mathbf{c}^\intercal_{\xi}\boldsymbol{\xi},
    \end{alignat}
\end{subequations}

where $\tilde{\delta}$ models the delayed steering angle set by the actuator. The time constant $T$ and damping ratio $D$ can be estimated by system identification methods. To embed the actuator dynamics into the MPC formulation, we extend the plant model of (\ref{linear_model}) by $\boldsymbol{\xi}$: 

\begin{equation}\label{state_space_extension}
\boldsymbol{\dot{\tilde{x}}}=\left[\begin{array}{c}
\boldsymbol{\dot{x}} \\
\boldsymbol{\dot{\xi}}
\end{array}\right]=\left[\begin{array}{cc}
\mathbf{A} & \mathbf{b} \mathbf{c^{\intercal}_{\xi}} \\
\mathbf{0} & \mathbf{A_{\xi}}
\end{array}\right]\left[\begin{array}{l}
\boldsymbol{x} \\
\boldsymbol{\xi}
\end{array}\right]+\left[\begin{array}{c}
\mathbf{0} \\
\mathbf{b_{\xi}}
\end{array}\right] u.
\end{equation}

Therefore, our MPC is able to perform predictions considering the steering delay of the actuator, thus providing more accurate sequences of control inputs and increasing the robustness of the controller against lateral delays.

\subsection{Active Yaw Control and Optimal Control Allocation}
\begin{figure}[H]
\centering
\definecolor{karaceingYellow}{rgb}{1,0.93,0.0}
\begin{tikzpicture}
\begin{axis}[
	legend cell align={left},
	legend style={fill opacity=0.8, draw opacity=1, text opacity=1, at={(0.98,0.02)}, anchor=south east, draw=white!80!black},
	axis equal image,
	grid = both,
		width=0.48*\textwidth,
	xlabel={$ a_{y} \ \mathrm{in} \ \frac{\mathrm{m}}{\mathrm{s}^2}$ },
    ylabel={$ a_{x} \ \mathrm{in} \ \frac{\mathrm{m}}{\mathrm{s}^2}$},
	legend cell align={left},
	enlargelimits=0.02,
	]
	\addplot[line width= 0.5mm, color=karaceingYellow,mark=*, only marks, mark options={scale=0.4}] table[x index=0,y index=1, col sep=comma] 
	{./plots/measured.csv};
	\addlegendentry{Meas.}
	\addplot[line width= 0.5mm, color=black,mark=*, only marks, mark options={scale=0.3}] table[x index=0,y index=1, col sep=comma] 
	{./plots/ref.csv};
	\addlegendentry{Ref.}
\end{axis}
\end{tikzpicture}
\caption{Accelerations during an Autocross run w/o prior track knowledge. Yellow dots: measured accelerations, black ellipse: target max. accelerations.}
\label{fig:gg_diag}
\end{figure}
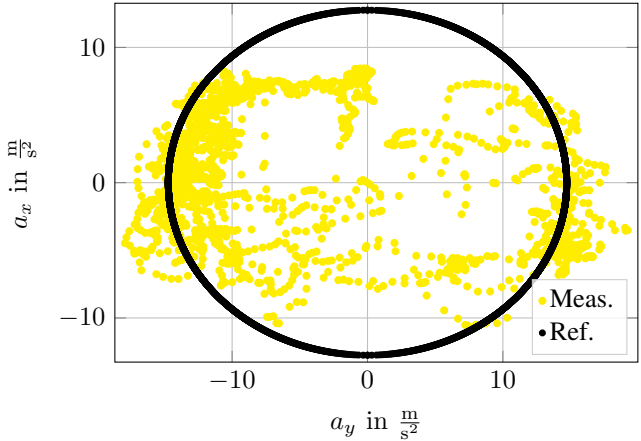

To utilize  the full potential of an all-wheel-drive electric race car, an optimal control algorithm was developed that maximizes grip at each wheel and actively controls the yaw rate of the car using differential torques. The underling optimal control problem is defined as follows:
\begin{equation}
    \min_{u} \left\|\boldsymbol{G} \boldsymbol{u}-\hat{\boldsymbol{u}}\right\|_{\mathbf{Q}}^{2}+ \left\|  \boldsymbol{u}\right\|_{\mathbf{R}}^{2}.
\end{equation}
The target control vector $\hat{\boldsymbol{u}} = [M_{z},F_x]^\intercal$ is composed of the desired yaw moment  and longitudinal force to track the reference velocity determined by the 2 DoF PI controller for velocity tracking. The desired yaw moment is calculated by a PID controller that uses the yaw rate predicted by the MPC for the first future state of the horizon as a reference input. This helps the vehicle behave more like the MPC assumes it will, and provides faster yaw rate buildup and stable handling.
The vector $ \boldsymbol{u} = [F_{fl}, F_{fr}, F_{rl}, F_{rr}]^\intercal $ represents the control vector, composed of the four individual driving forces.
The dependency matrix $\mathbf{G}$ describes the geometric relation between the driving forces and target  $\hat{\boldsymbol{u}} $ given by
\begin{equation}
   \mathbf{G}=\left[\begin{array}{cccc}
-s_f & s_f & -s_r & s_r \\
1 & 1 & 1 & 1 \\
\end{array}\right], 
\end{equation}
where $s_f, s_r$ are half the track width at the front and rear axles, respectively.
Each driving force in the control vector is penalized individually in the cost function, inversely proportional to the wheel load pressing on the tire. Thus, the wheel with the lowest wheel load or grip is penalized the most, resulting in that tire being used the least to drive the vehicle. The resulting vehicle control module allows stable handling even in the range of accelerations above 1g. The achieved accelerations during an autocross run are shown in Fig. \ref{fig:gg_diag}.
The obtained optimal control problem can be solved similarly to the MPC problem.

\section{Simulation}
For a continuous and hardware-independent software development, a precise vehicle dynamics simulation is needed. Because of our already existing lap time simulation for manually driven vehicles and to achieve a higher customizability, we decided to develop a simulation environment in-house. To avoid problems when transferring the software from simulation to the real car, we developed the entire vehicle dynamics simulation in ROS C++ with exactly the same interfaces as the real system.

To model the physics of our race car, we use a nonlinear dynamic 7 DoF planar vehicle model featuring a combined slip Pacejka tire model. The model also takes wheel load transfer, aerodynamic effects and the powertrain characteristics into account. In addition, virtual IMU signals and wheel odometry are added with sensor-typical noise. Further, delay time and latency of the actuator systems, especially of the steering system, are modelled to better adapt the simulation  to the real vehicle behavior. 
For validation, we compared simulation results with real vehicle data recorded on the same track outline, e.g. regarding the calculated and measured accelerations. A simulation of the perception was tested but later dropped in favor of using recorded data.
\comment{
\begin{figure}[h]
\begin{center}
\includegraphics[width=0.45\textwidth]{ADR_2021/images/SIMwithout.jpg}
\includegraphics[width=0.45\textwidth]{ADR_2021/images/simwith.jpg}
\caption{ Comparison of the path of the simulation with that of the real test with (bottom) and without (top) considering the actuator latency }\label{lactency}
\end{center}
\end{figure}
Figure \ref{lactency} shows how well the simulation models reality. This allows us to make good predictions of the lap time and optimize the parameters of the control system quite far without the need for real testing.}

In addition to the vehicle dynamic simulation, we evaluated a simulation of the perception pipeline in 2020. Though it was useful to virtually test different camera setups, we concluded that the perception is more efficiently tested on recorded data.
 
\section{Conclusion}

Using the autonomous software pipeline presented in this paper, we were able to win the overall competition at all attended events in 2021: in the Czech Republic, Hungary and Germany. We also achieved the fastest lap times at Formula Student Germany.

For the 2022 season, the team focuses on increasing software robustness and reliability as well as pushing the speed even further by evaluating approaches for non-linear control and online parameter estimation.

\printbibliography

\end{document}